# Better by You, better than Me? ChatGPT-3 as writing assistance in students' essays.

Željana Bašić, Ana Banovac*, Ivana Kružić, Ivan Jerković

University Department of Forensic Sciences, University of Split

Ruđera Boškovića 33, 21000 Split, Croatia

Correspondence to: Ana Banovac, ana.banovac@forenzika.unist.hr

**Abstract**

**Aim:** To compare students' essay writing performance with or without employing ChatGPT-3 as a writing assistant tool.

**Materials and methods:** Eighteen students participated in the study (nine in control and nine in the experimental group that used ChatGPT-3). We scored essay elements with grades (A-D) and corresponding numerical values (4-1). We compared essay scores to students' GPTs, writing time, authenticity, and content similarity.

**Results:** Average grade was C for both groups; for control (2.39 ± 0.71) and for experimental (2.00 ± 0.73). None of the predictors affected essay scores: group ($P = 0.184$), writing duration ($P = 0.669$), module ($P = 0.388$), and GPA ($P = 0.532$). The text unauthenticity was slightly higher in the experimental group (11.87% ±13.45 to 9.96% ± 9.81%), but the similarity among essays was generally low in the overall sample (the Jaccard similarity index ranging from 0 to 0.054). In the experimental group, AI classifier recognized more potential AI-generated texts.

**Conclusions:** This study found no evidence that using GPT as a writing tool improves essay quality since the control group outperformed the experimental group in most parameters.

## Introduction

November 30 2022, will go down in history as the date when a free version of the AI language model created by OpenAI called ChatGPT-3 [1] was made available for public usage. This language model's functions encompass text generation, answering questions, and completing tasks such as translation and summarization [2].

ChatGPT can be employed as assistance in the world of academia. It can improve writing skills since it is trained to deliver feedback on style, coherence, and grammar [3], extract key points, and provide citations [4]. This could increase the efficiency of researchers, allowing them to concentrate on more crucial activities (e.g., analysis and interpretation). This has been supported by studies showing that ChatGPT could generate abstracts [5,6], high-quality research papers [7], dissertations, and essays [3]. Previous studies showed that ChatGPT could create quality essays on different topics [8–13]. For example, this program, in conjunction with davinci-003, generated high-quality short-form essays on Physics, which would be awarded First Class, the highest grade in UK high education system [14]. It also led to questions on the ethics of using ChatGPT in different forms of academic writing, the AI authorship [7,15–18], and raised issues of evaluating academic tasks like students' essays [19–21]. Unavoidable content plagiarism issues were discussed, and solutions for adapting essay settings and guidelines were revised [8,14,20,22].

However, it is still unknown how ChatGPT performs in students' environment as a writing assistant tool and does it enhance students' performance. Thus, this research investigated whether ChatGPT would improve students' essay grades, reduce writing time, and affect text authenticity.

## Materials and methods

We invited the second-year master's students from the University Department of Forensic Sciences, University of Split, Croatia, to voluntarily participate in research on essay writing as a part of the course Forensic sciences seminar. Out of 50 students enrolled in the course, 18 applied by web form and participated in the study. Before the experiment, we divided them into two groups according to the study module (Crime Scene Investigation, Forensic Chemistry and Molecular Biology, and Forensics and National Security) and the weighted grade point average (GPA) to ensure the similar composition of the groups. The control group (n = 9, GPA = 3.92 ± 0.46) wrote the essay traditionally, while the experimental group (n = 9, GPA = 3.92 ± 0.57) used ChatGTP assistance, version 2.1.0. [1].

Before the study, the students signed the informed consent and were given a separate sheet to write their names and password. This enabled anonymity while grading essays and further analysis of student-specific variables. We explained the essay scoring methodology [23] to both groups, with written instructions about the essay title (The advantages and disadvantages of biometric identification in forensics sciences), length of the essay (800 – 1000 words in the Croatian language), formatting, and citation style (Vancouver). We introduced the experimental group to the ChatGPT tool. All students had four hours to finish the task and could leave whenever they wanted. The control group was additionally supervised to ensure they did not use the ChatGPT.

Two teachers graded the essays (ŽB, associate professor, and IJ, assistant professor). They compared the grades, and if their scoring differed the final grade was decided by the consensus. We used the essay rubrics from Schreyer Institute for Teaching Excellence, Pennsylvania State University, that included the following criteria (mechanics, style, content, and format) and grades from A to D [23]. We converted categorical grades to numbers (A=4, B=3, C=2, D=1) for further analysis. For each student, we recorded writing time.

We checked the authenticity of each document using PlagScan (PlagScan GmbH, Germany, 2023), and conducted the pairwise comparison for document similarity using R studio (ver. 1.2.5033) and package Textreuse [24] using the Jaccard similarity index. We checked the content using an AI text classifier to test if a human or an AI created the text. According to this classifier, text was scored as very unlikely, unlikely, unclear, possibly, and likely that it was AI generated [25]. We opted for this package after similar programs [25–27] did not recognize a ChatGPT-generated text in the Croatian language as AI-assisted text.

Statistical analysis and visualization were conducted using Excel (Microsoft Office ver. 2301) and R studio (ver. 1.2.5033). The final essay score was calculated as an average of four grading elements. The linear regression was used to test the effects of group, writing duration, module, and GPA on overall essay scores. The level of statistical significance was set at $P \leq 0.05$.

## Results

The duration of the essay writing for the GTP-assisted group was 172.22 ± 31.59, and for the control, 179.11 ± 31.93 minutes. GTP and control group, on average, obtained grade C, with a slightly higher average score in the control (2.39 ± 0.71) than in the GTP group (2.00 ± 0.73) (Figure 1A). The mean of text unauthenticity was 11.87% ±13.45 in the GPT-assisted group and 9.96% ± 9.81% in the control group. The text similarity in the overall sample was low (Supplementary table 1), with a median value of the Jaccard similarity index of 0.002 (0 – 0.054). The AI text classifier showed that, in the control group, two texts were possibly, one likely generated by AI, two were unlikely created by AI, and four cases were unclear. The ChatGPT group had three possible and five cases likely produced by AI, while one case was labeled as unclear.

Figure 1 (A and B) implies a positive association between duration and GPA with essay scores. However, students with higher GPAs in the control group achieved higher scores than those in the GTP group. The association of essay scores and non-authentic text proportion (Figure 1C) was detected only in the GPT group, where the students with more non-authentic text achieved lower essay scores.

The linear regression model showed a moderate positive relationship between the four predictors and the overall essay score ($R = 0.573$; $P = 0.237$). However, none of the predictors had a significant effect on the outcome: group ($P = 0.184$), writing duration ($P = 0.669$), module ($P = 0.388$), and GPA ($P = 0.532$).

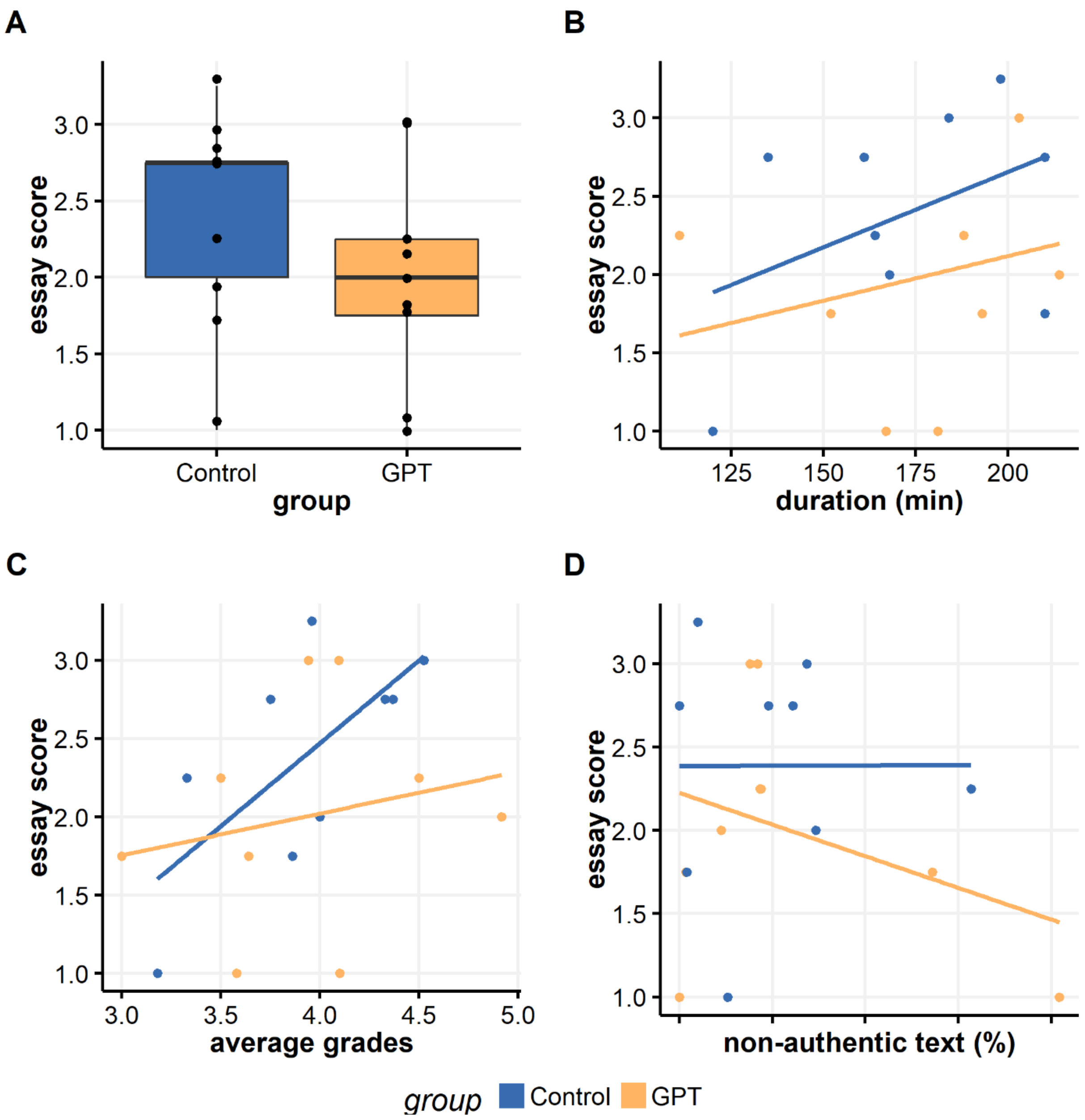

Figure 1. Essay scores by A) group, B) duration, C) average grades, D) proportion of non-authentic text.

Supplementary table 1. Pairwise comparison of essay similarity.

| | Control1 | Control2 | Control3 | Control4 | Control5 | Control6 | Control7 | Control8 | Control9 | GPT1 | GPT2 | GPT3 | GPT4 | GPT5 | GPT6 | GPT7 | GPT8 | GPT9 |
|---|---|---|---|---|---|---|---|---|---|---|---|---|---|---|---|---|---|---|
| **Control1** | *NA* | 0.0013 | 0.0000 | 0.0537 | 0.0014 | 0.0018 | 0.0041 | 0.0023 | 0.0044 | 0.0031 | 0.0015 | 0.0005 | 0.0000 | 0.0008 | 0.0022 | 0.0037 | 0.0049 | 0.0110 |
| **Control2** | *NA* | *NA* | 0.0015 | 0.0031 | 0.0030 | 0.0024 | 0.0031 | 0.0024 | 0.0010 | 0.0023 | 0.0026 | 0.0000 | 0.0010 | 0.0050 | 0.0005 | 0.0019 | 0.0014 | 0.0009 |
| **Control3** | *NA* | *NA* | *NA* | 0.0009 | 0.0011 | 0.0015 | 0.0021 | 0.0015 | 0.0000 | 0.0015 | 0.0017 | 0.0000 | 0.0010 | 0.0009 | 0.0000 | 0.0000 | 0.0000 | 0.0000 |
| **Control4** | *NA* | *NA* | *NA* | *NA* | 0.0051 | 0.0054 | 0.0099 | 0.0100 | 0.0014 | 0.0047 | 0.0044 | 0.0005 | 0.0022 | 0.0047 | 0.0013 | 0.0027 | 0.0004 | 0.0117 |
| **Control5** | *NA* | *NA* | *NA* | *NA* | *NA* | 0.0046 | 0.0054 | 0.0041 | 0.0033 | 0.0038 | 0.0017 | 0.0016 | 0.0051 | 0.0044 | 0.0005 | 0.0005 | 0.0044 | 0.0051 |
| **Control6** | *NA* | *NA* | *NA* | *NA* | *NA* | *NA* | 0.0117 | 0.0095 | 0.0021 | 0.0098 | 0.0022 | 0.0020 | 0.0015 | 0.0017 | 0.0005 | 0.0005 | 0.0005 | 0.0074 |
| **Control7** | *NA* | *NA* | *NA* | *NA* | *NA* | *NA* | *NA* | 0.0118 | 0.0057 | 0.0128 | 0.0044 | 0.0000 | 0.0013 | 0.0011 | 0.0006 | 0.0013 | 0.0019 | 0.0104 |
| **Control8** | *NA* | *NA* | *NA* | *NA* | *NA* | *NA* | *NA* | *NA* | 0.0021 | 0.0085 | 0.0060 | 0.0010 | 0.0010 | 0.0034 | 0.0009 | 0.0035 | 0.0005 | 0.0074 |
| **Control9** | *NA* | *NA* | *NA* | *NA* | *NA* | *NA* | *NA* | *NA* | *NA* | 0.0020 | 0.0006 | 0.0000 | 0.0000 | 0.0027 | 0.0020 | 0.0021 | 0.0031 | 0.0042 |
| **GPT1** | *NA* | *NA* | *NA* | *NA* | *NA* | *NA* | *NA* | *NA* | *NA* | *NA* | 0.0051 | 0.0019 | 0.0037 | 0.0028 | 0.0018 | 0.0028 | 0.0027 | 0.0112 |
| **GPT2** | *NA* | *NA* | *NA* | *NA* | *NA* | *NA* | *NA* | *NA* | *NA* | *NA* | *NA* | 0.0022 | 0.0054 | 0.0121 | 0.0010 | 0.0093 | 0.0057 | 0.0043 |
| **GPT3** | *NA* | *NA* | *NA* | *NA* | *NA* | *NA* | *NA* | *NA* | *NA* | *NA* | *NA* | *NA* | 0.0030 | 0.0030 | 0.0014 | 0.0025 | 0.0020 | 0.0035 |
| **GPT4** | *NA* | *NA* | *NA* | *NA* | *NA* | *NA* | *NA* | *NA* | *NA* | *NA* | *NA* | *NA* | *NA* | 0.0090 | 0.0000 | 0.0055 | 0.0077 | 0.0015 |
| **GPT5** | *NA* | *NA* | *NA* | *NA* | *NA* | *NA* | *NA* | *NA* | *NA* | *NA* | *NA* | *NA* | *NA* | *NA* | 0.0000 | 0.0060 | 0.0100 | 0.0030 |
| **GPT6** | *NA* | *NA* | *NA* | *NA* | *NA* | *NA* | *NA* | *NA* | *NA* | *NA* | *NA* | *NA* | *NA* | *NA* | *NA* | 0.0028 | 0.0041 | 0.0019 |
| **GPT7** | *NA* | *NA* | *NA* | *NA* | *NA* | *NA* | *NA* | *NA* | *NA* | *NA* | *NA* | *NA* | *NA* | *NA* | *NA* | *NA* | 0.0068 | 0.0085 |
| **GPT8** | *NA* | *NA* | *NA* | *NA* | *NA* | *NA* | *NA* | *NA* | *NA* | *NA* | *NA* | *NA* | *NA* | *NA* | *NA* | *NA* | *NA* | 0.0062 |
| **GPT9** | *NA* | *NA* | *NA* | *NA* | *NA* | *NA* | *NA* | *NA* | *NA* | *NA* | *NA* | *NA* | *NA* | *NA* | *NA* | *NA* | *NA* | *NA* |

## Discussion

As we are aware, this is the first study that tested ChatGPT-3 as an essay-writing assistance tool in a student population sample. It showed that the aid of ChatGPT did not necessarily improve the quality of students' essays. The ChatGPT group did not perform better in either of the indicators; the students did not deliver higher quality content, did not write faster, nor had a higher degree of authentic text.

The overall essay score was slightly better in the control group, which could probably result from the over-reliance on the tool or students' unfamiliarity with it. This was in line with Fyfe's study on writing students' essays using GPT-2, where students reported that it was harder to write using the tool than by themselves. Students also raised the question of not knowing the sources of generated text which additionally distracted them in writing task [28]. Some studies did show more promising results [8–14], but unlike our study, they were mainly based on GPT and experienced researcher interaction. This could be a reason for the lower performance of our GTP group, as the experienced researchers are more skilled in formulating questions, guiding the program to obtain better-quality information, and critically evaluating the content.

The other interesting finding is that the use of ChatGPT did not accelerate essay writing and that the students of both groups required a similar amount of time to complete the task. As expected, the longer writing time in both groups related to the better essay score. This finding could also be explained by students' feedback from Fyfe's study, where they specifically reported difficulties combining the generated text and their own style [29]. So, although ChatGPT could accelerate writing in the first phase, it requires more time to finalize the task and assemble content.

Our experimental group had slightly more problems with plagiarism than the control group. Fyfe also showed that his students felt uncomfortable writing and submitting the task since they felt they were cheating and plagiarizing [29]. However, a pairwise comparison of essays in our study did not reveal remarkable similarities, indicating that students had different reasoning and style, regardless of whether they were using ChatGPT. This could also imply that applying the tool for writing assistance produces different outcomes for the same task, depending on the user's input [14].

The available ChatGPT text detector [25] did not perform well, giving false positive results in the control group. Most classifiers are intended for English and usually have disclaimers for performance in other languages. This raises the necessity of improving existing algorithms for different languages or developing language-specific ones.

The main concern of using GPT in academic writing has been the unauthenticity [8,14,22], but we believe that such tools will not increase the non-originality of the published content or students' assignments. The detectors of AI-generated text are developing daily, and it is only a matter of time before highly reliable tools are available. When we consider the perspectives of detection tools and our findings where the students with GPT assistance did not outperform the control group, we can see no reason for a major concern about its application in academic writing.

The main drawback of this study is the limited sample size which does not permit the generalization of the findings or a more comprehensive statistical approach. One of the limitations could also be language-specificity (our students wrote in Croatian for their convenience), which disabled us from the full application of AI detection tools. We should also consider that ChatGPT is predominantly fed with English content, so we cannot exclude the possibility that writing in English could have generated higher-quality information. Lastly, this was our students' first interaction with ChatGPT, so it is possible that lack of experience also affected their performance. Future studies should therefore expand the sample size, number, and conditions of experiments, include students of different profiles, and extend the number of variables that could generally relate to writing skills.

As it seems, the academia and media concern about this tool might be unjustified, as, in our example, the ChatGPT was found to perform similarly to any web-based search: the more you know – the more you will find. In some ways, instead of providing structure and facilitating writing, it could distract students and make them underperform.